\documentclass[10pt,twocolumn,letterpaper]{article}

\usepackage{cvpr}
\usepackage{times}
\usepackage{epsfig}
\usepackage{graphicx}
\usepackage{amsmath}
\usepackage{amssymb}
\usepackage{array}
\usepackage{multirow}
\newcolumntype{H}{>{\setbox0=\hbox\bgroup}c<{\egroup}@{}}

\newcommand{\minisection}[1]{\vspace{0.04in} \noindent {\bf #1}\ \ }

\usepackage[pagebackref=true,breaklinks=true,colorlinks,bookmarks=false]{hyperref}

\cvprfinalcopy 

\def\cvprPaperID{****} 

\ifcvprfinal\fi
\begin{document}
\title{Visual attention models for scene text recognition}

\author{Suman K. Ghosh\\
Computer Vision Center\\
Barcelona, Spain\\
{\tt\small sghosh@cvc.uab.es}
\and
Ernest Valveny\\
Computer Vision Center\\
Barcelona, Spain\\
{\tt\small ernest@cvc.uab.es}
\and
Andrew D. Bagdanov\\
Media Integration and
Communication Center (MICC)\\ Universita di Firenze, Firenze\\
bagdanov@cvc.uab.es.
{\tt\small bagdanov@cvc.uab.es}}

\maketitle

\begin{abstract}
  In this paper we propose an approach to lexicon-free recognition of
  text in scene images. Our approach relies on a LSTM-based soft
  visual attention model learned from convolutional features. A set of
  feature vectors are derived from an intermediate convolutional layer
  corresponding to different areas of the image. This permits encoding
  of spatial information into the image representation. In this way,
  the framework is able to learn how to selectively focus on different
  parts of the image. At every time step the recognizer emits one
  character using a weighted combination of the convolutional feature
  vectors according to the learned attention model. Training can be
  done end-to-end using only word level annotations. In addition, we
  show that modifying the beam search algorithm by integrating an
  explicit language model leads to significantly better recognition
  results. We validate the performance of our approach on standard
  SVT and ICDAR'03 scene text datasets, showing state-of-the-art performance in unconstrained text recognition.
\end{abstract}


\section{Introduction}
The increasing ability to capture images in any condition and
situation poses many challenges and opportunities for extracting
visual information from images. One such challenge is the detection
and recognition of text ``in the wild''. Text in natural images is
high level semantic information that can aid automatic image
understanding and retrieval.

Text can also play a role in a number of applications such as
automatic translation, assisting the visually impaired, and robot
navigation \cite{case2011autonomous}.
However, robust reading of text in
uncontrolled environments is challenging due to a multitude of factors
such as difficult acquisition conditions, low resolution, font
variability, complex backgrounds, different lighting conditions, blur,
etc.
%
%
%
\begin{figure}
\centerline{\includegraphics[width=0.9\columnwidth]{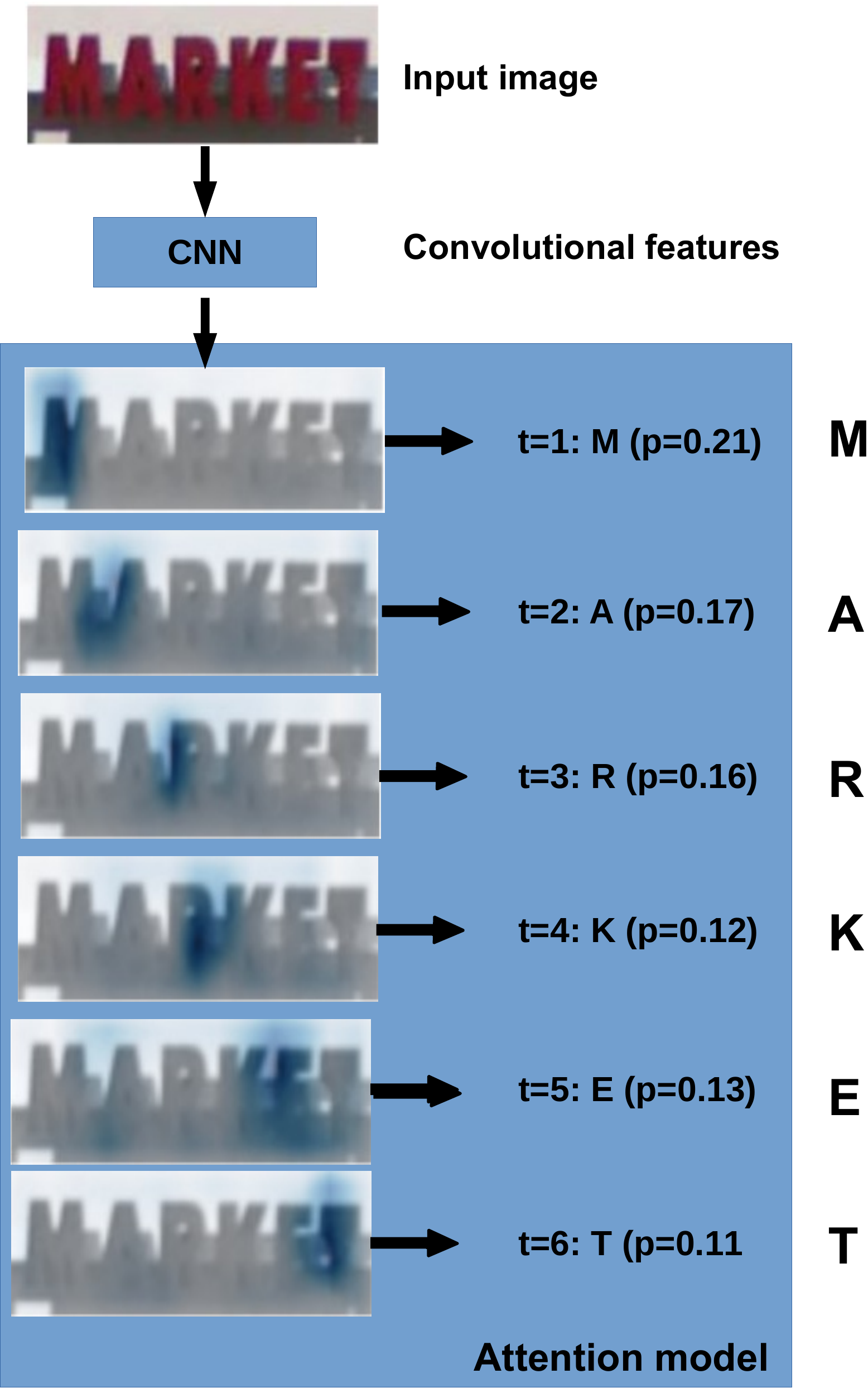}}
\label{fig:abs}
\caption{Overall scheme of the proposed recognition framework. Given a cropped word image, a set of spatially localized features are obtained using a CNN. Then, an LSTM decoder is combined with an attention model to generate the sequence of characters. At every time step the attention model weights the set of feature vectors to make the LSTM focus on a specific part of the image.}
\end{figure}
Recognizing text in outdoor scene images is very different than text
recognition in document images, where the problem is well studied and
many commercially successful Optical Character Recognition (OCR)
systems exist. OCR techniques used in document images do not
generalize to recognition of scene text because they are tuned for
document images having black text on white backgrounds and are mostly
scanned using flatbed scanner with uniform lighting and exposure.



The problem of end-to-end scene text recognition is usually divided in
two different tasks: word detection and word recognition. The goal of
the word detection stage is to generate bounding boxes around
potential words in the images. Subsequently, the words in these
bounding boxes are recognized in the word recognition stage.  This
paper is focused on this second stage, word recognition.

%
%
%


Many existing scene text recognition methods rely on a predefined
lexicon or dictionary~\cite{almazan2014word,Gordo_2015_CVPR,jaderberg2014deep_b,neumann2012real,Rodriguez-Serrano2015,smith2011enforcing,wang2011end}, which greatly improves accuracy by
constraining output. Although the use of a predefined lexicon
restricts the set of possible words to be recognized, excellent
results have been reported recently with dictionaries as large as
90,000 words~\cite{jaderberg2015}. Lexicon-based recognition works because the
lexicon acts as language prior or a recognition context. However, in
many cases such contextual information is either unavailable or hard
to determine. For instance, in case of images containing proper nouns
like names of brands or products, street names, etc. To deal with such
problems unconstrained text recognition methods have also been
proposed~\cite{photoocr,jaderberg2014deep_a,DBLP:journals/corr/LeeO16} (here by \emph{unconstrained} we mean that any word
can be recognized). In this work we focus on unconstrained text
recognition.





A class of methods for scene text
recognition~\cite{photoocr,wang2011end} use an over-segmentation to
generate multiple hypotheses of character locations, and then follow
with a supervised character classifier. Word recognition is the result
of finding the best sequence of character hypotheses according to
classifier scores and a set of spatial or lexical constraints.  These
methods fit the unconstrained scenario quite naturally, but can be
limited by the performance of character segmentation which is a
challenging step due to the nature of scene text. Alternatively,
another set of methods use a fixed-length representation to represent
word images holistically and avoid the segmentation step. Then this
representation is used either to classify words images among 1 of $k$
predefined word classes~\cite{jaderberg2014deep_b} or rank words from
a fixed lexicon~\cite{almazan2014word,Rodriguez-Serrano2015} based on
some sort of distance criteria. These approaches are mostly linked to
lexicon-based word recognition although in some cases they have also
been applied to unconstrained text recognition by inferring characters
and $n$-grams from the fixed-length
representation~\cite{jaderberg2014deep_a}.  However, this increases
the training complexity and requires incremental training and
heuristic gradient rescaling.

In contrast to the above strategies our approach neither recognizes
individual characters in the word image nor uses any holistic
representation to recognize the word. It rather uses a LSTM-based
visual attention model (based on \cite{DBLP:journals/corr/XuBKCCSZB15}) to focus attention on
relevant parts of the image at every step and infer a character
present in the image. Thus, the system is able to recognize
out-of-vocabulary words, although it does not need explicit character
segmentation or recognition. The visual attention model can be trained
using only word bounding boxes and does not need explicit character
bounding boxes at training time. As the model relies on recurrent
neural networks, it also learns an implicit language model from the
data, which is crucial in cases such as those shown in
\cite{photoocr}. However, the model also has the flexibility to
integrate an explicit language model, which it is shown in the
experiments to improve accuracy. Additionally, the output can be
constrained to a fixed lexicon (when available) to work in a
dictionary-based setting. An overview of the proposed recognition framework is illustrated in figure \ref{fig:abs}. 

The rest of the paper is organized as follows. In
Section~\ref{sec:related_work} we analyze the works most related to our
proposed approach. Then, we present our attention-based recognition approach
in Sections~\ref{sec:LSTM} and \ref{sec:beam_search}. In Section~\ref{sec:exp} we
experimentally validate the model on a variety of standard and public
benchmark datasets. We conclude in Section~\ref{sec:conclusion} with a
summary of our contributions and a discussion of future research
directions.

\section{Related Work}
\label{sec:related_work}

In this section we briefly review work from the literature on robust scene text recognition and attention-based
recognition models most related to our approach.

\minisection{Dictionary-based scene text recognition.}  Traditionally,
scene text recognition systems use character recognizers in a
sequential way by localizing characters using a sliding
window~\cite{jaderberg2014deep_b,neumann2012real,wang2011end} and then
grouping responses by arranging the character windows from left to
right as words. A variety of techniques have been used to classify
character bounding boxes, including random ferns~\cite{wang2011end},
integer programming~\cite{smith2011enforcing} and Convolutional Neural
Networks (CNNs)~\cite{jaderberg2014deep_b}. These methods often use
the lexical constrains imposed by a fixed lexicon while grouping the
character hypotheses into words.

In contrast to sequential character recognizer models, holistic
fixed-length representations have been proposed
in~\cite{almazan2014word,Gordo_2015_CVPR,Rodriguez-Serrano2015}. These
works advocate the use of a joint embedding space between images and
words. In addition, Gordo \etal in \cite{Gordo_2015_CVPR} make use of
supervised mid-level feature learned using character bounding boxes to
further improve the image-text embedding. In contrast, Yao \etal in
\cite{Yao_2014_CVPR}, used mid-level features which can be learned from the data in
a unsupervised way.
 
More recently, with the success of deep-CNN features in computer
vision, convolutional features have also been applied to scene text
recognition. The first such attempt was made by Jaderberg \etal
in~\cite{jaderberg2014deep_b}, where a sliding window over CNN features
is used for robust scene text recognition using a fixed lexicon. Later, the same authors also proposed a fixed-length representation~\cite{jaderberg2015} using convolutional features trained of a synthetic dataset of 9 million images~\cite{jaderberg2014synthetic}
 
\minisection{Unconstrained scene text recognition.}  Though most of
the works in scene text recognition focus on fixed-lexicon
recognition, a few attempts at unconstrained text recognition have
also been made. 
 
Biassco \etal in \cite{photoocr} rely on sequential character
classifiers.  They use a massive number of annotated character
bounding boxes to learn character classifiers. Binarization and
sliding window methods are used to generate character proposals
followed by a text/background classifier. Finally, character
probabilities given by character classifiers are used in a beam search
to recognize words. They also integrate a static character $n$-gram
language model in every step of the beam search to incorporate an
underlying language model.
 
Though CNN models have achieved great success in lexicon-based text
recognition, word recognition in unconstrained scenarios requires
modeling the underlying character-level language model.  Jaderberg
\etal in~\cite{jaderberg2014deep_a} proposed to use two separate CNNs,
one modeling character unigram sequences and another $n$-gram language
statistics. They additionally use a Conditional Random Field to model
the interdependence of characters ($n$-grams). However, this
significantly increases computational complexity.  In addition, to
detect the presence of character $n$-grams in word images as neural
activations, character $n$-grams are used as output nodes, leading to a
huge (10k output units for $n$=4) output layer.

To model inter-dependencies between characters, the authors
of~\cite{DBLP:journals/corr/LeeO16} used a recursive CNN and variants
of Recurrent Neural Networks on top of CNN features. 



 
\minisection{Visual attention models for recognition.}
Recently visual attention models have gained a lot of attention and
have been used for machine
translation~\cite{DBLP:journals/corr/BahdanauCB14} and image
captioning~\cite{DBLP:journals/corr/XuBKCCSZB15}. In this last work the attention
model is combined with an LSTM on top of CNN features. The LSTM outputs one word at
every step focusing on a specific part of the image driven by the attention model. Two models 
of attention, hard and soft attention are proposed. In our work, we mainly follow the
soft attention model, adapted to the particular case of text recognition. Attention models appear to have the potential ability to overcome some of the limitations of existing text recognition methods. They can leverage a fixed-length representation, but at the same time, they are able to guide recognition to relevant parts of the image, performing in this way a kind of implicit character segmentation.

Recently, a soft attention model has also been proposed for text recognition in the wild~\cite{DBLP:journals/corr/LeeO16}. 
The main differences with our work are the following. Firstly, the success of~\cite{DBLP:journals/corr/LeeO16} can be largely attributed to
the use of Recursive Neural Network (RNN) features. 
They rely on the RNN features to model the dependencies between characters. Instead we use
traditional CNN features and it is the visual attention model who learns to
selectively attend to parts of the image and the dependencies between
them. Secondly, Lee \etal \cite{DBLP:journals/corr/LeeO16} used the
features from fully connected layer, while we use features from an
earlier convolutional layer, thus preserving the local spatial
characteristics of the image and reducing the model complexity. 
This also allows the model to focus on a
subset of features corresponding to certain area of the image and
learn the underlying inter-dependencies.  Thirdly, we used LSTM
instead of RNN which has been shown to learn long term dependencies
better than traditional RNNs.

\minisection{Our contributions with respect to the state-of-the-art.}
In summary the contributions of our work are:
\begin{itemize}
\item We introduce a LSTM-based visual attention model for
  unconstrained scene text recognition. This model is able to
  selectively attend to specific parts of word images, allowing it to
  model inter-character dependencies as needed and thus to
  \emph{implicitly} model the underlying language.
\item We show that weak explicit language models (in the form of prefix
  probabilities) can significantly boost the final recognition result
  without having to resort to a fixed lexicon. For that, We modify the beam
  search to take into account the language model. Additionaly, the beam search 
  can also incorporate a lexicon whenever it is available.
\item We experimentally validate that our approach
    with weak language modeling outperforms the state-of-the-art in
    unconstrained scene text recognition and performs comparably to
    lexicon-based approaches with a model complexity lower than similar 
    approaches.
\end{itemize}
\section{Visual attention for scene text recognition}

\label{sec:LSTM}
\begin{figure}
\includegraphics[scale=1]{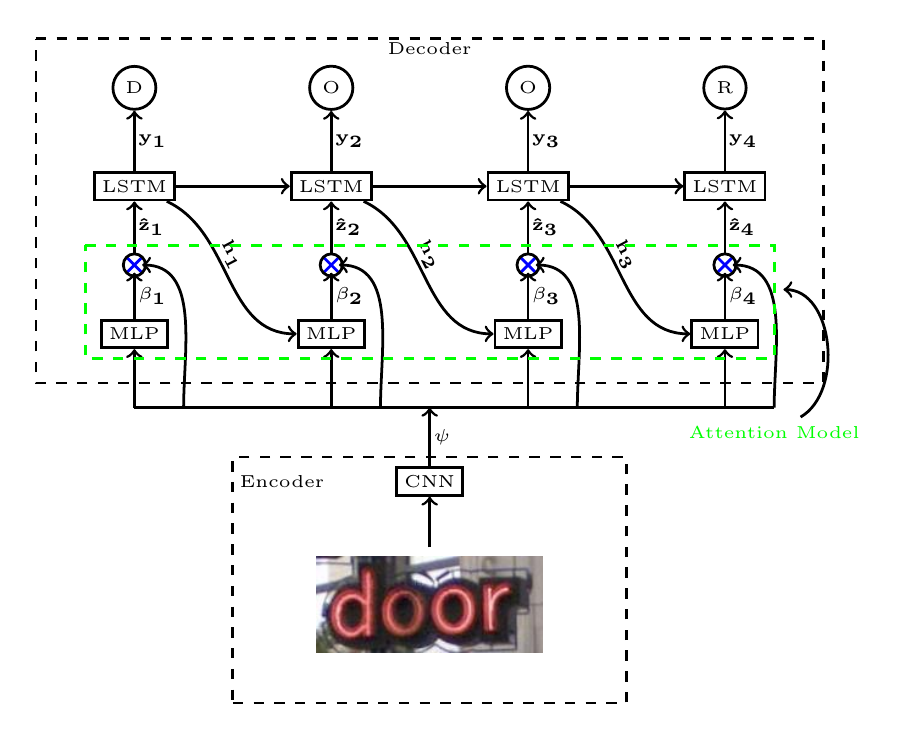}
\caption{The proposed Encoder-decoder framework with attention model.}
\label{fig:LSTM}
\end{figure}

Our recognition approach is based on an encoder-decoder framework for
sequence to sequence learning. An overall scheme of the framework is
illustrated in figure~\ref{fig:LSTM}. The encoder takes an image of a
cropped word as input and encodes this image as a sequence of
convolutional features. The attention model in between the encoder and
the decoder drives, at every step, the focus of attention of the
decoder towards a specific part of the sequence of features. Then, an
LSTM-based decoder generates a sequence of alpha-numeric symbols as
output, one at every time step, terminating when a special stop symbol
is output by the LSTM. Below we describe the details of each of the
components of the framework.

\minisection{Encoder:}
The encoder uses a convolutional neural network to extract a set of
features from the image. Specifically, we make use of the CNN model
proposed by Jaderberg et al.~\cite{jaderberg2015} for scene text
recognition -- however we do not use the fully connected layer as a
fixed-length representation as it is common in previuos
works. Instead, we take the features produced by the last
convolutional layer. In this way we can produce a set of feature
vectors, each of them linked to a specific spatial location of the
image through its corresponding receptive field.
This preserves
spatial information about the image and reduces model
complexity. Through the attention model, the decoder is able to use
this spatial information to selectively focus on the most relevant
parts of the image at every step.

Thus, given an input image of a cropped word, the encoder generates a
set of feature vectors:
\begin{eqnarray}
\Psi= \{x_i: i=1 \ldots K\},
\end{eqnarray}
where $x_i$ denotes the feature vector corresponding to $i^{th}$ part
of the image. Each $x_i$ corresponds to a spatial location in the
image and contains the activations of all feature maps at that
location in the last convolutional layer of the CNN.

\minisection{Attention model:} For the attention model, we adapt the
soft attention model of~\cite{DBLP:journals/corr/XuBKCCSZB15}
for~image captioning, originallly introduced
by~\cite{DBLP:journals/corr/BahdanauCB14} for neural machine
translation. In~\cite{DBLP:journals/corr/XuBKCCSZB15} slightly better
results are obtained using the hard version of the model that focuses,
at every time step, on a single feature vector.  However, we argue
that, in the case of text recognition, the soft version is more
appropriate since a single character will usually span more than one
spatial cell of the image corresponding to each of the feature
vectors. The soft version of the model can combine several feature
vectors with different weights into the final representation.

As shown in figure \ref{fig:LSTM}, the attention model generates,
at every time step $t$, a vector $\hat{z_t}$ that will is the input to
the LSTM decoder. This vector $\hat{z_t}$ can be expressed as a
weighted combination of the set $\Psi$ of feature vectors $x_i$
extracted from the image:
\begin{equation}
\hat{z_t} = \sum^K_{i=1}{\beta_{t,i}x_i}
\end{equation}

Thus, the vector $\hat{z_t}$ encodes the relative importance of
each part of the image in order to predict the next character for the
underlying word.  At every time step $t$, and for each location $i$ a
positive weight $\beta_{t,i}$ is assigned such that
$\sum(\beta_i)=1$. These weights are obtained as the softmax
output of a Multi Layer Perpectron (denoted as $\Phi$) using the set
of feature vectors $\Psi$ and the hidden state of the LSTM decoder at
the previous time step, $h_{t-1}$. More formally:
\begin{eqnarray}
\alpha_{ti} &=& \Phi\left(x_i,h_{t-1}\right) \\
\beta_{ti} &=& \frac{\exp\left(\alpha_{ti}\right)}{\sum^K_{j=1}\exp\left(\alpha_{t,j}\right)}
\end{eqnarray}
This model is smooth and differentiable and thus it can be learned
using standard back propagation.

\minisection{Decoder:} Our decoder is a Long Short Term Memory (LSTM)
network~\cite{hochreiter1997long} which produces one symbol from the
given symbol set $L$, at every time step.  The output of the LSTM is a
vector $y_t$ of $|L|$ character probabilities which represents the
probability of emitting each of the characters in the symbol set $L$
at time $t$. It depends on the output vector of the soft attention
model $\hat{z_t}$, the hidden state at previous step $h_{t-1}$ and the
output of the LSTM at previous step $y_{t-1}$. We follow the notation
introduced in~\cite{DBLP:journals/corr/XuBKCCSZB15} where the network is described
by:
\begin{eqnarray}
\left( \begin{array}{c} i_t \\ f_t\\o_t\\g_t \end{array} \right) &=& \left( \begin{array}{c} \sigma \\ \sigma\\\sigma\\\tanh \end{array} \right) T \left( \begin{array}{c} Ey_{t-1} \\ h_{t-1}\\ \hat{z_t}\\ \end{array} \right)\\
c_t &=& f_t \odot c_{t-1} + i_t \odot g_t\\
h_t &=& o_t \odot \tanh \left(c_t\right),
\end{eqnarray}
where $T$ is the matrix of weights learned by the network and
$i_t, f_t , c_t , o_t,$ and $ h_t$ are the input, forget, memory,
output and hidden state of the LSTM, respectively. In the above
definition, $\odot$ denotes the element-wise multiplication and $E$ is an
embedding of the output character probabilities that is also learned
by the network. $\sigma$ and $\tanh$ denote the activation functions
that are applied after the multiplication by the matrix of weights

Finally, to compute the output character probability $y_t$, a deep
output layer is added that takes as input the character probability at
the previous step, the current LSTM hidden state, and the current
feature vector. The output character probability is:
\begin{equation}
P\left(y_t\middle| \Psi,y_{t-1}\right) \sim \exp\left(\mathtt{L_0}\left( Ey_{t-1} + \mathtt{L_h}h_t+\mathtt{L_z}\hat{z_t} \right) \right)
\end{equation}
where 
$\mathtt{L_0}$, $\mathtt{L_h}$ and $\mathtt{L_z}$ are the parameters of the deep output layer that are learned using back-propagation.

\section{Inference}
\label{sec:beam_search}
We use beam search over LSTM outputs to perform word inference. We
first introduce the basic procedure, and then describe how we extend
it to incorporate language models.

\subsection{The basic inference procedure}
Once the model is trained, we use a beam search to approximately
maximize the following score function over every possible word:
$\mathtt{w} = [c_1, \dots, c_n]$:
\begin{equation}
  \label{eqn:score}
  S\left(\mathtt{w},x\right) = \sum_{t=1}^N \log \left(P\left(c_t\middle| c_{t-1}\right)\right),
\end{equation}
where $c_n$ is a special symbol signifying the end of a word, which
immediately stops the beam search.
 
The beam search keeps track at every step of the top $N$ most probable
sequences of characters. For every active branch of the beam search,
given the previous character of the sequence, $c_{t-1}$, the output
character probability $y_t$ of the LSTM is used to obtain
$P\left(c_t\middle| c_{t-1}\right)$ for all characters $c_t$ in the
symbol set $L$.

\subsection{Incorporating language models}
\label{sec:language_model}
Text is a strongly contextual. There are some strict constraints
imposed by the grammar of the language. For example any word in
English cannot carry more than two consecutive occurrences of any
alphabet letter. Leveraging such knowledge can positively impact the
final recognition output. Although the LSTM implicitly learns some
dependences between consecutive characters, we show that adding an
explicit language model that takes into account longer dependencies
gives a significant boost to recognition accuracy.
 
In this work we use a standard $n$-gram based language model during
inference to leverage the language prior. The character $n$-gram model
gives probability of a character conditioned on $k$ previous
characters, where $k$ is a parameter of the model:
\begin{equation}
\Theta\left(c_k\middle| c_{k-1},c_{k-2}...,c_1 \right) = \frac{\#\left(c_1 c_2...c_{k-1}\right)}{\#\left(c_k c_k ... c_k\right)},
\end{equation}
where, $\#(c_1, \dots\ c_n)$ is the number of occurrences of a
particular substring in a training corpus.
 
Finally, the score function in equation \ref{eqn:score} can be
modified to take the $n$-gram language model into account as:
\begin{eqnarray}
  S\left(\mathtt{w},x\right) & = & \sum_{t=1}^N \log \left(P\left(c_t\middle| c_{t-1}\right)\right) \nonumber \\
                             & + & \alpha \log \Theta\left(w_t\middle| w_{t-1},w_{t-2}...,w_1 \right)
                             \label{eqn:beamsearch}
\end{eqnarray}
 
At every step we fix the parameter $k$ of the language model to the
number of previously generated characters in order to take into
account the longest possible sequence.

\subsection{Lexicon-based inference}
\label{sec:lexicon_based}
Although our method is originally designed for unconstrained text
recognition, it can also leverage a lexicon whenever available.  The
use of a lexicon $D$ can be integrated by modifying the beam search so
that all active sequences that do not correspond to any valid word are
automatically removed from the beam. This can be expressed by
modifying equation \ref{eqn:score} as:
\begin{equation}
    S\left(\mathtt{w},x\right) =
    \begin{cases}
      \begin{aligned}
        \sum_{t=1}^N & \log \left(P\left(c_t\middle| c_{t-1}\right)\right)\\
        & + \alpha \log \Theta\left(w_t\middle| w_{t-1},\ldots,w_1 \right) 
      \end{aligned}
      & \mbox{if } \mathtt{w} \in $D$ \\
      -\infty & \mbox{if} \mathtt{w} \notin D
    \end{cases}    
\end{equation}
 
This can be efficiently implemented by storing the lexicon in a trie structure and automatically removing from the beam search any alternative that do not correspond to any partial branch of the trie.

\section{Experimental Results}
\label{sec:exp}
In this section we report on experiments carried out to validate the
proposed model for unconstrained scene text recognition.

\subsection{Datasets and experimental protocols}
We evaluate the performance of the proposed method using the following
standard datasets.

\minisection{Street View Text (SVT) dataset:} this dataset contains
647 cropped word images downloaded from Google Street View. In
addition to results on totally unconstrained recognition we also
report results using the predefined lexicons defined by Wang \etal
in~\cite{wang2011end}. Results with the 50-word lexicon are referred
as SVT-50.

\minisection{ICDAR'03 text dataset:} this dataset dataset contains 251
full images and 860 cropped word images~\cite{icdar2003}. We used the
same protocol as
\cite{almazan2014word,DBLP:journals/corr/LeeO16,wang2011end} and
evaluate cropped word images for which the groundtruth text contains
only alphanumeric characters and contains at least three characters.
As before we report results on both unconstrained and lexicon-based
recognition scenarios. We used same strategy as Wang \etal
\cite{wang2011end} to create a 50-word lexicon (ICDAR'03-50), while
the setup using the lexicon of all ground truth words is referred as
ICDAR'03-full.

\minisection{Synth90k text dataset:} this dataset is used only for
training~\cite{jaderberg2014synthetic}. It contains 9 million
synthetically-generated text images. We use the official partition for training as in other works like \cite{DBLP:journals/corr/LeeO16}.


\minisection{Evaluation protocol:} We use the standard evaluation
protocol adopted in most previous work on text recognition in scene
images~\cite{jaderberg2014deep_a,DBLP:journals/corr/LeeO16,wang2011end}.
The accepted metric is word level accuracy in percentage. SVT and
ICDAR'03 are used for evaluation.  For lexicon-based recognition, we
used the same set of 50 for all images in for SVT and ICDAR'03
dataset, as proposed buy Wang \etal~\cite{wang2011end}.

\minisection{Implementation details:}
The CNN encoder used in this work is the Dictnet model by Jaderberg \etal \cite{jaderberg2014synthetic}. Their  deep convolutional network consists of four convolutional layers and two fully connected layers. In this work we used features from the last convolutional layer. Thus, the feature map used is of size $4\times 13 \time 512$ and therefore, the LSTM takes input in the form of $52 \times 512$. 

For lexicon-based recognition when we do not use the lexicon-based
inference explained in section~\ref{sec:lexicon_based}. Instead, we
take the output of unconstrained recognition and find the closest word
in the lexicon using the Levenshtein edit distance. For lexicon-based
inference in unsconstrained datasets (SVT and ICDAR'03) we use the
90k-words lexicon provided by Jaderberg \etal in
\cite{jaderberg2014synthetic}. The explicit language model is also
learned using this 90k word lexicon. 

The parameter $\alpha$ (see equation \ref{eqn:beamsearch}) to weight the language model with respect to LSTM character probability is empirically established. In our experiments we found the best results with $\alpha$ between 0.25 to 0.3
\subsection{Baseline performance analysis}
In this section we analyze the impact on performance of all the
components of the proposed model. We start with a baseline that
consists of a simple one layer LSTM network as decoder, without any
attention or explicit language model. As we are interested mainly in
the impact of the attention model, we use a simple version in which
CNN features from the encoder are fed to the LSTM only at the first
time step. At every step the output character is determined based on
the output of the previous step and the previous hidden state.

In an effort to evaluate each of our contributions, we trained the
baseline system and our model with exactly the same training data. For
this purpose we randomly sampled one million training samples from the
Synth90k~\cite{jaderberg2014synthetic} dataset. For validation we used
300,000 samples randomly taken from the same synth90K dataset.

We present the results for each of the component of the framework as
described above in Table~\ref{tab:baseline_study}. The attention model
outperforms the baseline by a significant margin (around 7\%). Also
these results confirm the advantage of using an explicit language
model in addition to the implicit conditional character probabilities
learned by the LSTM model. Using the language model improves accuracy
in another 7\%. We also see that further constraining the inference
wih a dictionary does not improve the result much, probably because
the language model is learned from the same 90K dictionary proposed by
Jaderberg \etal in~\cite{jaderberg2014synthetic}.

In comparison with other related works on unconstrained text
recognition, it is noteworthy that with only one million training
samples our complete framework can learn a better model than Jaderberg
\etal~\cite{jaderberg2014deep_a} and obtain results that are close to
other state-of-the-art methods that are using the whole 9 million sample
training dataset (see table~\ref{table:comp_pref}).


\begin{table}[h]
\centering
\begin{tabular}{l|c}
\hline
\textbf{Methods} & \textbf{SVT} \\
 \hline
Baseline (LSTM-no attention) &61.7 \\
Proposed (LSTM + attention model) &68.16 \\
Proposed (LSTM + attention model + LM)&75.57\\ 
Proposed (LSTM + attention model+LM+dict)&76.04\\ 
\hline
\end{tabular}
\caption{Impact of the different components of our framework with respect to the baseline. We compare the baseline (LSTM with no attention model) with all the variants of the proposed method, incrementally: using only the attention model (section \ref{sec:LSTM}, integrating also the explicit language model (section \ref{sec:language_model}, and contraining the inference to a lexicon (section \ref{sec:lexicon_based}).
} 
\label{tab:baseline_study}
\end{table}

\begin{table*}[ht]
\centering
\small
\begin{tabular}{l|l|c|c|H H H|c|c|c H}
\hline
 &
\textbf{Methods} & \textbf{SVT-50} & \textbf{SVT} & \textbf{IIIT5k-50} &
\textbf{IIIT5k-1k} & \textbf{IIIT5k} & \textbf{ICDAR'03-50} &
\textbf{ICDAR'03-full}& \textbf{ICDAR'03}&\textbf{ICDAR'13}\\
\hline
\parbox[t]{2mm}{\multirow{15}{*}{\rotatebox[origin=c]{90}{\textbf{Lexicon-based recognition}}}}
& ABBY Baseline ABBYY \cite{wang2011end}&35.0&-&24.3&-&-&56.0&55.0&-&-\\
& Wang \etal\cite{wang2011end}&57.0&-&-&-&-&76.0&62.0&-&-\\
& Mishra \etal \cite{Mishra2012}&73.2&-&-&-&-&81.8&67.8&-&-\\
& Novikova \etal \cite{Novikova2012}&72.9&-&64.1&57.5&-&82.8&-&-&-\\
& Wang \etal \cite{Wang2012}&70.0&-&-&-&-&90.0&84.0&-&-\\
& Goel \etal \cite{Goel2013}&77.3&-&-&-&-&89.7&-&-&-\\
& Alsharif and Pineau \cite{Alsharif2013EndtoEndTR}&74.3&-&-&-&-&93.1&88.6&-&-\\
& Almazan \etal \cite{almazan2014word}&89.2&-&91.2&82.1&-&-&-&-&-\\
& Lee \etal \cite{Lee2010} & 80.0&-&-&-&-&88.0&76.0&-&-\\
& Yao \etal \cite{Yao_2014_CVPR}&75.9&-&80.2&69.3&-&88.5&80.3&-&-\\
& Rodriguez-Serrano \etal\cite{Rodriguez-Serrano2015}&70.0&-&76.1&57.4&-&-&-&-&-\\
& Jaderberg \etal \cite{jaderberg2015}&86.1&-&-&-&-&96.2&91.5&-&-\\
& Su and Lu\etal \cite{}&83.0&-&-&-&-&92.0&82.0&-&-\\
& Gordo \etal \cite{Gordo_2015_CVPR}&90.7&-&93.3&86.6&-&-&-&-&-\\
& *DICT Jaderberg et al. \cite{jaderberg2014synthetic}&95.4&80.7&97.1&92.7&-&98.7&98.6&93.1&90.8\\ \hline
\parbox[t]{2mm}{\multirow{6}{*}{\rotatebox[origin=c]{90}{\textbf{Unconstrained}}}}
& Bissacco \etal \cite{photoocr}&90.4&78.0&-&-&-&-&-&-&87.6\\
& Jaderberget al. ~\cite{jaderberg2014deep_a}&93.2&71.7&95.5&89.6&-&97.8&97.0&89.6&81.\\
& Lee \etal \cite{DBLP:journals/corr/LeeO16}&96.3&80.7&96.8&94.4&78.4&97.9&97.0&88.7&90.0\\
& Proposed (LSTM + attention model) &91.7&75.1&&&&93.4&91.0&89.3&\\
& Proposed (LSTM + attention model + LM)&95.2&80.4&&&&95.7&94.1&92.6&\\
& Proposed (LSTM + attention model+LM+dict)&95.4&-&-&&&96.2&95.7&-&\\
\hline
\end{tabular}
\caption{Scene text recognition accuracy. ``50''and ``Full'' denote the lexicon size used for constrained text
  recognition as defined in~\cite{wang2011end}. Results are divided
  into lexicon-based and unconstrained (lexicon-free) approaches. *DICT \cite{jaderberg2014synthetic}
  is not lexicon-free due to incorporating ground-truth labels during training.}
\label{table:comp_pref}
\end{table*}

\subsection{Comparison with state of the art}
In this section we will compare our result with other related works on scene text recognition. The results of this comparison are shown in table~\ref{table:comp_pref}. First, we will discuss results on unconstrained text recognition which is the main focus of our work. Then, we will analyze results for lexicon-based recognition. 

\minisection{Unconstrained text recogntion: }apart from our method Jaderberg \etal \cite{jaderberg2014deep_a}, Lee \etal \cite{DBLP:journals/corr/LeeO16} and Bissaccco \etal \cite{photoocr} are the only methods which are capable of performing totally unconstrained recognition of scene text. Among these methods, our visual attention based model performs significantly better than Bissacco \etal \cite{photoocr} and Jaderberg \etal \cite{jaderberg2014deep_a} in both SVT and ICDAR'03 datasets. Our model also performs as good as Lee \etal \cite{DBLP:journals/corr/LeeO16} in SVT dataset and outperforms them by 3\% in ICDAR'03 dataset, which is significant given the high recognition rates. 


If we further compare our model with that of Lee \etal \cite{DBLP:journals/corr/LeeO16}, that also uses different variants of RNN architectures and an attention model on top of CNN features, we find that they use recursive CNN features. They report that this gives an 8\% increase in accuracy over the baseline. This success is due to the recurrent nature of the CNN feature which implicitly model the conditional probability of character sequences.using recursive CNN performs better than the traditional convolutional feature. However, the RNN architecture they use improves only 4\% over the baseline. In contrast our method rely on traditional CNN features (which can possibly encodes the presence of individual characters as shown in \cite{jaderberg2014deep_a} from lower convolutional layer preserving local spatial characteristics, which reduces the complexity of the model. In addition, as reported in table~\ref{tab:baseline_study}, our combination of LSTM and soft attention model achieves a much larger margin, 14\%, over the baseline. Theses results show that a combination of local convolutional features using the context based attention attention performs better or comparable to the previous state-of-the- art results.


\minisection{Lexicon-based recognition}
For SVT-50 we can observe that our method obtain a similar result than the best of the methods~\cite{jaderberg2014synthetic} specifically designed to work in a lexicon-based scenario. Comparing with methods for unsconstrained text recognition, only the method of Lee\etal~\cite{DBLP:journals/corr/LeeO16} outperforms our best setting. But as we have already discussed, part of this better performance can be explained by the use of the more complex recursive CNN features.

Concerning ICDAR'03-50 and ICDAR'03-full, our results, although do not beat current state of the art are very competitive and comparable to the best performing methods.

\subsection{Discussion}

\begin{figure}
\includegraphics[scale=0.9]{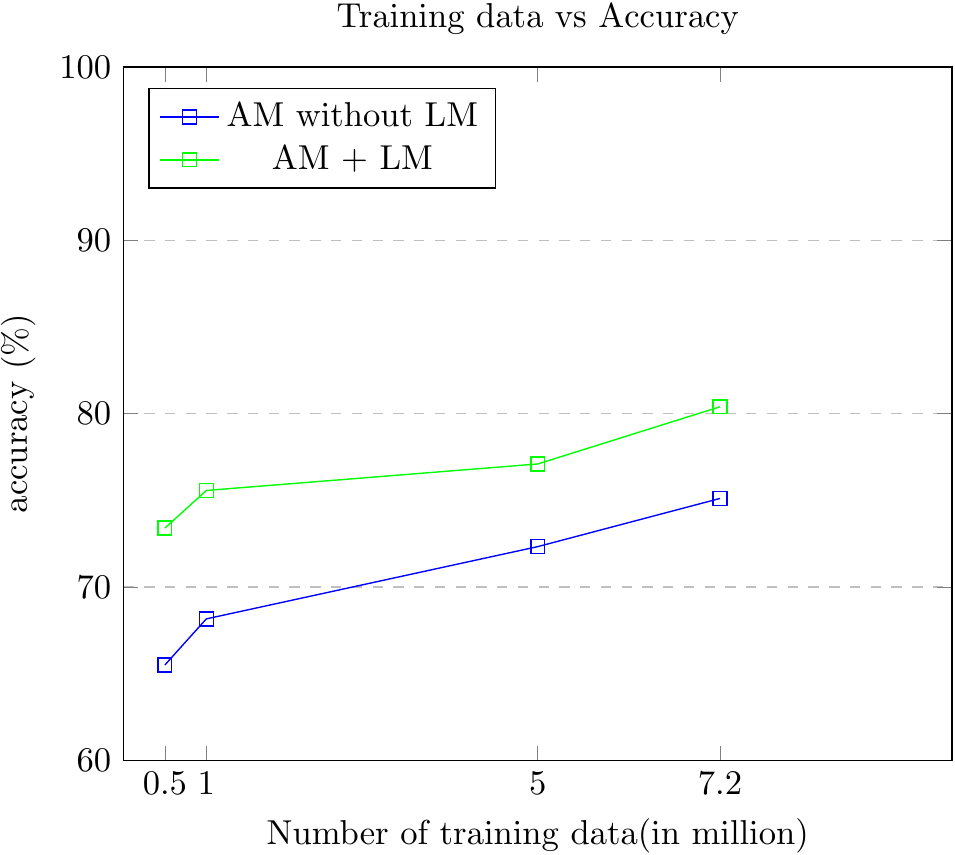}
\caption{Effect of increasing size of the training set data with
  (green) and without the language model (blue).}
\label{fig:training_data}
\end{figure}

\begin{figure*}
\includegraphics[scale=0.73]{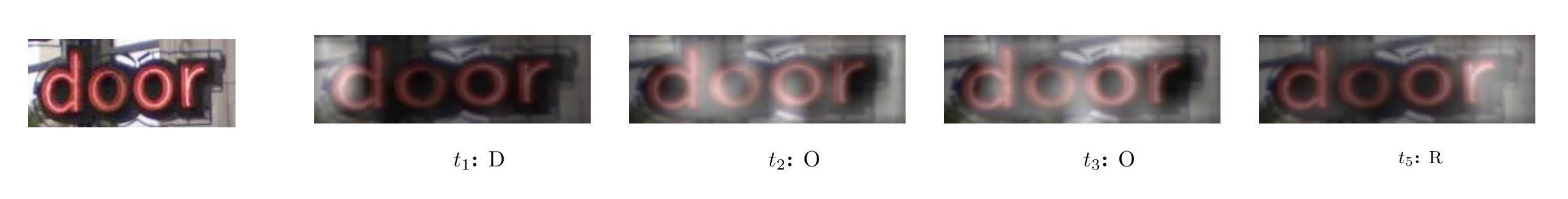}
\includegraphics[scale=0.93]{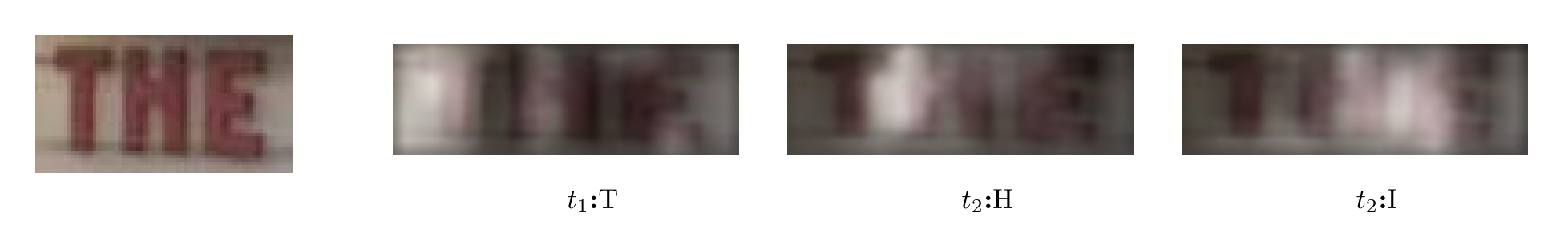}
\includegraphics[scale=0.45]{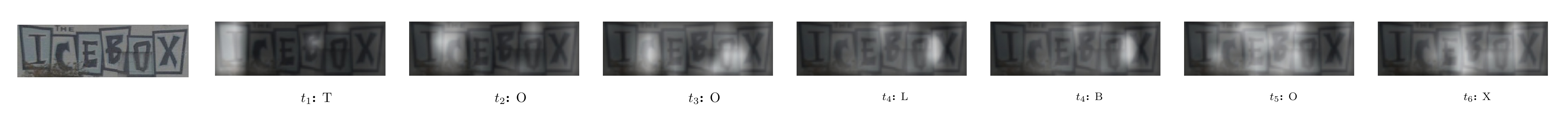}
\caption{Some results obtained with the proposed model.}
\label{fig:qualitative_results}
\end{figure*}

\minisection{Effect of language modeling:} The LSTM with attention model implicitly learns some character sequence model as the output at every step is conditioned by the previous character and also the attention is conditioned by the previous hidden state. However, the  baseline analysis in \ref{tab:baseline_study} shows that apart from this implicit character sequence model, adding an explicit language model helps to improve the performance. This finding is inline with the works of Bissacco \etal \cite{photoocr}, who also uses a static n-gram language model learned by a huge in-house corpus.
In our case we used the 90k dictionary as corpus to learn the language probability. To efficiently calculate the language probability we make use of a trie \cite{Fredkin:1960:TM:367390.367400} data structure. In Bissacco \etal, authors used a n-gram model with $n=8$ and additionally they also used word probabilities to re-rank the words, for which they learn additional word level language model. The authors analyzed the impact of the language model and they report a reduction of the word error rate by approximately 39.2\%. In contrast, our language model is small and improves the recognition accuracy around 4\%, a result which is very consistent across the different datasets -- see Table~\ref{table:comp_pref}.

\minisection{Effect of Training Data}
As the famous quote of Google’s Research Director Peter Norvig ''We do not have better algorithms. We just have more data.''. The effectiveness of data can not be denied in today's machine learning systems. As the real datasets are too small for learning deep networks, Jaderberg \etal \cite{jaderberg2014synthetic} proposed to learn deep models using only a synthetic dataset. This dataset has 9 million cropped word images. In order to analyze the behavior of our attention models with increasing data, we plot in Figure \ref{fig:training_data} the recognition accuracy in SVT dataset with respect to the number of training samples. As expected, accuracy increased with the training size. 

We also analyzed the effect of language model with an increasing number of training samples. Initially with 0.5 million training data the improvement when adding the language model is around 8\%. However as we increase the training data size this reliance on language model reduces to around 4\%. One reason for this can be that with more data the implicit language model learned by the LSTM is more powerful. The use of an LSTM capable of modeling character sequences can also be an additional reason that explains why the improvement obtained by Bissacco \etal using the language model is greater. 


\minisection{Qualitative Results}
As the features used to encode the image correspond to different parts of the image, we can visualize the attention at every time step of the network. In Figure \ref{fig:qualitative_results} we can see those visualizations.
We can observe that every time step a part of the image is attended, in a way that roughly mimics the natural reading order. We can also notice that the attention model is performing some kind of implicit character segmentation.

\section{Conclusions}
\label{sec:conclusion}

In this paper we proposed an LSTM-based visual attention model for
scene text recognition. The model uses convolutional features from a
standard CNN as input to an LSTM network that selectively attends to
parts of the image at each time step in order to recognize words
without resorting to a fixed lexicon. We also propose a modified beam
search strategy that is able to incorporate weak language models
($n$-grams) to improve recognition accuracy. Experimental results
demonstrate that our approach outperforms or performs
comparably to state-of-the-art approaches that use lexicons to
constrain inferred output words. Experimental results shows that context plays a important part in case of real data, thus using a explicit language model always helps to improve the result. 

In future we can extend the attention model for the text detection task, which will lead to an end-to-end framework for text recognition from images. Moreover, in our current framework convolutional features are taken from one single layer, which can lead to poorer results when the text is either too big or too small. This can be dealt with combining features from multiple layers.

{\small
\bibliographystyle{ieee}
\bibliography{egbib}
}

\end{document}